\documentclass{article}
\usepackage{ijcai23}

\usepackage{times}
\usepackage{soul}
\usepackage{url}
\usepackage[hidelinks]{hyperref}
\usepackage[utf8]{inputenc}
\usepackage[small]{caption}
\usepackage{graphicx}
\usepackage{amsmath}
\usepackage{amsthm}
\usepackage{amssymb}
\usepackage{booktabs}
\usepackage{algorithm}
\usepackage[switch]{lineno}
\usepackage{multicol}
\usepackage{multirow}
\usepackage{makecell}
\usepackage{diagbox}
\usepackage{subfigure}
\usepackage[noend]{algpseudocode}
\usepackage{svg}
\usepackage{array}
\usepackage[export]{adjustbox}
\usepackage{bm}
\usepackage{amsfonts}
\usepackage{hhline}
\graphicspath{ {fig/} }

\title{SeRO: Self-Supervised Reinforcement Learning for Recovery from Out-of-Distribution Situations}

\author{
Chan Kim$^1$
\and
Jaekyung Cho$^1$\and
Christophe Bobda$^2$\and
Seung-Woo Seo$^1$\And
Seong-Woo Kim$^1$
\affiliations
$^1$Seoul National University\\
$^2$University of Florida
\emails
\{chan\_kim, jackyoung96, sseo, snwoo\}@snu.ac.kr,
cbobda@ece.ufl.edu
}
\begin{document}

\maketitle

\begin{abstract}
    Robotic agents trained using reinforcement learning have the problem of taking unreliable actions in an out-of-distribution (OOD) state. Agents can easily become OOD in real-world environments because it is almost impossible for them to visit and learn the entire state space during training. Unfortunately, unreliable actions do not ensure that agents perform their original tasks successfully. Therefore, agents should be able to recognize whether they are in OOD states and learn how to return to the learned state distribution rather than continue to take unreliable actions. In this study, we propose a novel method for retraining agents to recover from OOD situations in a self-supervised manner when they fall into OOD states. Our in-depth experimental results demonstrate that our method substantially improves the agent’s ability to recover from OOD situations in terms of sample efficiency and restoration of the performance for the original tasks. Moreover, we show that our method can retrain the agent to recover from OOD situations even when in-distribution states are difficult to visit through exploration. Code and supplementary materials are available at \href{https://github.com/SNUChanKim/SeRO}{https://github.com/SNUChanKim/SeRO}.
\end{abstract}

\section{Introduction}

\label{sec:introduction}

Reinforcement learning (RL) has been used to solve challenging tasks in the field of robotics control and has achieved human-level performance \cite{schulman2015high,heess2017emergence,gu2017deep,akkaya2019solving}. However, several limitations prevent RL from being applied in real-world environments. One of the main limitations is the unreliable actions of RL agents in out-of-distribution (OOD) states that deviate from the learned state distribution. While operating in real-world environments, agents can fall easily into OOD states because the state space is extensive and non-stationary, which makes it impossible for the agent to cover the entire space during training. Unfortunately, unreliable actions in OOD states can lead to the failure of the agent \cite{amodei2016concrete} because they do not ensure that the agent performs its original tasks successfully. Therefore, agents should learn how to return to learned state distribution from the time they recognize that they have fallen into an OOD state, rather than continue to take unreliable actions. Take a quadruped walking robot trained using RL as an example and suppose that the robot has never been overturned during training. If the robot collides with a person outside the robot’s field of view while operating in the real world, it can overturn and unintentionally fall into an OOD state. In this situation, unreliable actions will not enable the robot to operate in its original purpose (\textit{walking}) because the robot has never been trained for such a situation. Instead of taking unreliable actions, the robot should learn how to \textit{turn its body over}, which enables it to return to the learned state distribution. However, the desired behavior for returning to the learned state distribution differs depending on the environment and the OOD situation. Hence, designing the corresponding reward function for each environment and OOD situation is laborious and requires prior knowledge.

Several studies in model-based RL have proposed methods to prevent agents from falling into OOD situations \cite{kahn2017uncertaintyaware,lutjens2019safe,henaff2019model,kang2022lyapunov}. However, these methods focus on discouraging agents from visiting OOD states. They do not consider the situations where trained agents have already fallen into OOD states, which can unintentionally occur in the real world.

In this study, we propose an RL method for \textbf{Se}lf-supervised \textbf{R}ecovery from \textbf{O}OD situations (SeRO), which can retrain agents to recover from OOD situations without prior knowledge of the environment or OOD situations. Recovery from OOD situations involves both 1) learning to return to the learned state distribution from the OOD situations and 2) restoring performance for the original task after the return. Unlike previous studies that aim to \textit{prevent} the agent from falling into OOD situations, our method aims to \textit{recover} the agent when the trained agent has already fallen into an OOD state during operation. We emphasize that our method is orthogonal and complementary to previous studies that focus on preventing agents from falling into OOD situations. 

We denote the agent's training for solving the original task as the \textit{training phase}, whereas the \textit{retraining phase} refers to the additional training required for returning to the learned state distribution and restoring original performance when the trained agent falls into OOD states during operation. When the agent is in OOD states, the agent cannot know the reward function for the desired behavior of returning to the learned state distribution without prior knowledge of the environment and OOD situation. To retrain the agent to return without an explicit reward designed based on prior knowledge, we propose an intrinsically motivated auxiliary reward that increases as the agent approaches the learned state distribution. The auxiliary reward is implemented based on a metric called the uncertainty distance, which we introduce to approximate the relative distance of the state from the learned state distribution. However, learning the behaviors to return to the learned state distribution can be thought of as learning a new task that is different from the original task. This can cause the agent to forget the original task while learning to return to the learned state distribution. To prevent such a situation, we propose uncertainty-aware policy consolidation. The main contributions of our paper can be summarized as follows:

\begin{itemize}
\item We propose a self-supervised RL method, SeRO, which retrains the agent to recover from OOD situations in a self-supervised manner.
\item We introduce a metric called the uncertainty distance, which approximately represents the relative distance of the state from the learned state distribution.
\item Our in-depth experimental results demonstrate that our method substantially improves the agent’s ability to recover from OOD situations in terms of sample efficiency and restoration of the performance for the original tasks.
\item Moreover, we demonstrate that the proposed method can successfully retrain the agent to recover from OOD situations even when in-distribution states are difficult to visit through exploration.
\end{itemize}

\section{Related Work}

\label{sec:citations}
\subsection{Preventing OOD Situations in RL}

Several attempts have been made to prevent an agent from falling into OOD situations in model-based RL. Kahn \textit{et al.} \shortcite{kahn2017uncertaintyaware} and Lütjens \textit{et al.} \shortcite{lutjens2019safe} proposed methods that used model-based RL with model predictive control (MPC) to prevent agents from falling into uncertain situations such as collision. In these works, the uncertainty of the learned model is calculated for all motion primitives and MPC chooses the action sequence among motion primitives by considering their cost and uncertainty. Similarly, Henaff \textit{et al.} \shortcite{henaff2019model} proposed a method that uses the uncertainty of the learned model to regularize the agent to stay in in-distribution states. Instead of using predefined motion primitives, they penalize the policy during learning when the simulated trajectory generated by the policy causes high uncertainty in the learned model. Kang \textit{et al.} \shortcite{kang2022lyapunov} suggested combining concepts from the density model that estimates training data distribution and the Lyapunov stability \cite{Sastry1999} to avoid distribution shifts when using learning-based control algorithms.

In the field of offline RL \cite{levine2020offline}, the policy is trained using a fixed offline dataset without additional interaction with the environment. Offline RL methods are more likely to fall into OOD situations because the distribution of the offline dataset cannot entirely cover that of the environment. To address this problem, Fujimoto \textit{et al.} \shortcite{pmlr-v97-fujimoto19a} and Kumar \textit{et al.} \shortcite{NEURIPS2019_c2073ffa} proposed methods to regularize the policy towards the distribution of an offline dataset, and Wu \textit{et al.} \shortcite{wu2019behavior}, Kumar \textit{et al.} \shortcite{NEURIPS2020_0d2b2061}, and Li \textit{et al.} \shortcite{pmlr-v164-li22d} proposed methods to penalize the value function when the policy deviates from the distribution of an offline dataset during training.

However, these methods focus on preventing the agent from falling into OOD situations by discouraging the policy from selecting the action that leads the agent to OOD states. Unlike previous works, our method addresses situations where the agent has already fallen into OOD situations unexpectedly during operation. Because previous works focus on preventing OOD situations, they do not deal with such situations. In this study, we focus on retraining the agent to recover from such situations in a self-supervised manner, which has the concept of \textit{recovery} rather than \textit{prevention}. Therefore, our method is orthogonal and complementary to previous works that focus on preventing OOD situations.

\subsection{Returning to a Particular State Distribution in RL}
In the field of autonomous RL \cite{sharma2022autonomous} which aims to train the agent in non-episodic environments, the agent is trained to autonomously return to states to restart the training. Eysenbach \textit{et al.} \shortcite{eysenbach2017leave} proposed a method to train a reset policy that leads the agent to a \textit{predefined} initial state distribution. However, in this method, the reward for the training reset policy must be defined based on prior knowledge of the initial state distribution, e.g., in locomotion environments, the reset reward is large when the agent is standing upright. Alternatively, in order to accelerate learning, Sharma \textit{et al.} suggested a training policy to return to states selected by a value-based curriculum \shortcite{sharma2021autonomous} or the distribution of expert demonstrations \shortcite{sharma2022state}, rather than a predefined initial state. In the field of safe RL, Thananjeyan \textit{et al.} \shortcite{thananjeyan2021recovery} suggested a training recovery policy to return to safe states. In this study, a safety critic that indicates the risk of the action in the given state is \textit{pretrained} using an offline dataset that contains controlled demonstrations of constraint-violating behavior. During policy training, when the policy generates an action with high risk, a recovery policy is executed to generate an action to return to a state with low risk.

These methods all address returning from states that are included in training environments while our method addresses returning from OOD states that are completely \textit{unseen in the training environments}. Moreover, these methods only deal with tasks where returning to a particular state distribution is symmetric to original tasks, e.g., in navigation tasks, \textit{moving to particular positions} for returning is symmetric to \textit{moving to goal positions}, and in manipulation tasks, \textit{collocating objects in particular positions} for returning is symmetric to \textit{collocating objects in goal positions}. Conversely, our method deals with tasks where returning is not symmetric to original tasks, e.g., the agent should learn to \textit{turn its body over} to return to the learned state distribution, whereas the original task is \textit{moving forward as fast as possible}.

\section{Preliminaries}
\label{sec:prelimnaries}
\subsection{Uncertainty Estimation}
Uncertainty can be categorized into two types: aleatoric and epistemic \cite{review}. Aleatoric uncertainty, also known as data uncertainty, is caused by the inherent randomness of an input. In contrast, epistemic uncertainty, also known as model uncertainty, is caused by insufficient knowledge about the data. In this study, we focus on epistemic uncertainty and refer to it as \textit{uncertainty} throughout the paper for the sake of simplicity. Uncertainty can be represented as the variance of the posterior distribution of network parameters $W$. Hence, knowledge about the posterior distribution of parameters $p(W|D)$ is required, where $D$ refers to the data composed of input $x$ and output $y$. However, the posterior distribution does not exist in a general neural network because a general neural network is optimized over the deterministic parameter. On the other hand, in Bayesian neural networks (BNN) \cite{Goan_2020}, $p(W|D)$ exists by assuming the prior distribution of the parameter $p(W)$. In BNN, given a deterministic input $x$, the output $y$ is a random variable because the parameter $W$ is assumed to be a random variable, and the distribution of $y$ given $D$ and $x$ is as follows:
\begin{equation}
\footnotesize
p(y|D, x) = \int_W{p(y|x, W)p(W|D) \,dW},
\end{equation}
When data that the network has never learned is given, the variance of the posterior distribution $p(W|D)$ increases, thereby increasing the variance of output distribution $p(y|D,x)$. This enables the estimation of uncertainty $\sigma^u$ via the variance of output distribution as $\sigma^{u} = \textrm{Var}_{p(W|D)}(y)$.

Gal \textit{et al.} \shortcite{mcdo} proposed a Monte Carlo dropout (MCD) method that uses dropout as a Bayesian approximation to estimate uncertainty. They proved that using multiple forward passes of the network with random dropout activation approximates the Bayesian inference of the deep Gaussian process. The uncertainty estimated using MCD for neural network $f$ with input $x$ is calculated as follows:
\begin{equation}
\label{eqn:mcdo_cal}
\footnotesize
\sigma^{u} \approx \sum_{i=1}^{N}{\frac{(f^i(x)-{\bar{f}})^2}{N}},\ \textrm{with}\ \bar{f} = \sum_{i=1}^{N}\frac{f^i(x)}{N},
\end{equation}
where $N$ is the number of forward passes, and $f^i$ refers to the $i^{\textrm{th}}$ forward pass of neural network $f$ with dropout activation. We used MCD to estimate the uncertainty of the state, as it is empirically proven to accurately represent the uncertainty in RL \cite{pmlr-v139-wu21i}.

\subsection{Soft Actor-Critic (SAC)}
SAC \cite{sac} is an off-policy actor-critic method based on a maximum entropy RL framework, which aims to maximize expected reward while also maximizing entropy. Improvement in exploration and robustness due to maximum entropy formulation enables SAC to solve the challenges of high sample complexity and brittle convergence properties of model-free RL. In this method, soft value iteration for updating the Q-function and the policy is proposed. Soft value iteration alternates soft policy evaluation for updating the Q-function and soft policy improvement for updating the policy. It is proven that the policy provably converges to the optimal policy through soft policy iteration. The objectives for updating the parameterized Q-function $Q_\theta$ and the policy $\pi_\phi$ are as follows:
\begin{equation}
\label{policy_eval}
\footnotesize
J(\theta)=\mathbb{E}_{s_t,a_t}\left[\frac{1}{2}\left(Q_\theta(s_t,a_t)-y\right)^2\right],
\end{equation}
\begin{equation}
\label{policy_improv}
\footnotesize
    J(\phi)=\mathbb{E}_{s_t,a_t}\left[\alpha\log\pi_\phi(a_t|s_t)-Q_\theta(s_t,a_t)\right],
\end{equation}
with {\footnotesize$y=r_t+\gamma\mathbb{E}_{s_{t+1}, a_{t+1}}\left[Q_\theta(s_{t+1}, a_{t+1})-\alpha\log\pi_\phi(a_{t+1}|s_{t+1}))\right]$}, where $\alpha$ is entropy coefficient, $\theta$ and $\phi$ are the parameters of the Q-function and the policy respectively. We refer the readers to the original paper for proof of convergence.
\section{Method}

\label{sec:proposed_method}
In this section, we introduce the SeRO framework, which aims to retrain agents to recover from OOD situations in a self-supervised manner. SeRO is implemented by expanding SAC. Note that although our method expands SAC, it can be combined with any RL algorithm trained using the policy gradient method. We propose a novel auxiliary reward that increases as the agent approaches the learned state distribution. When the agent is in the OOD state during the retraining phase, it is trained using the auxiliary reward until it returns to the in-distribution state. Moreover, in order to prevent the agent from forgetting the original task while learning to return to the learned state distribution, we used uncertainty-aware policy consolidation. In the remainder of this section, we describe how our method is implemented in detail.
\begin{figure}[t]
    \centering
    \includegraphics[width=\linewidth]{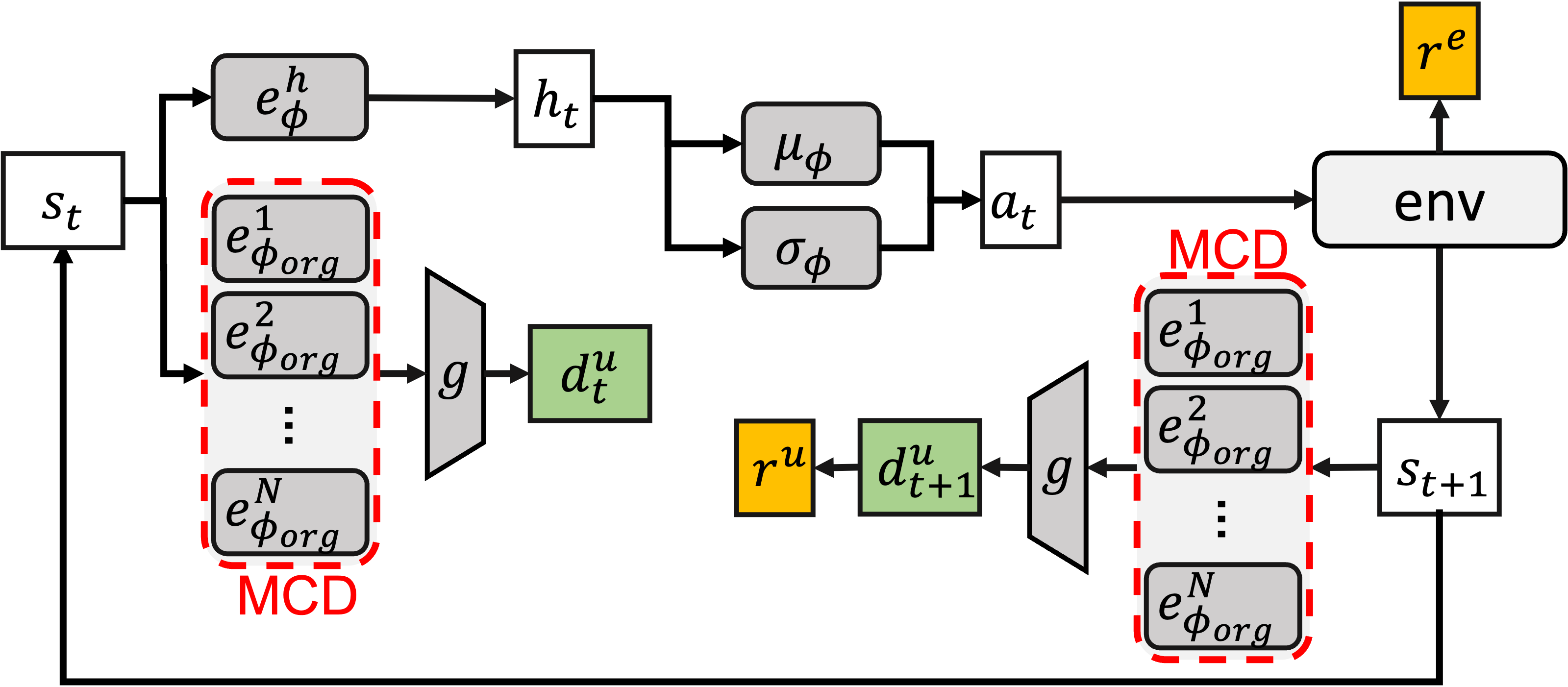}
    \caption{Environment step of SeRO in the retraining phase: All networks with subscript $\phi$ are the components of the policy $\pi_\phi$. $e_\phi^n$ with $n=h,1,2,..., N$ refers to the encoder network $e_\phi$ with dropout activation, and $\mu_\phi$ and $\sigma_\phi$ refer to the networks that generate the parameters of the Gaussian action distribution. Note that, the uncertainty distance $d^u_t$ is calculated using the fixed original policy $\pi_{\phi_{org}}$.}
    \label{fig:arch}
    \vspace{-1.2em}
\end{figure}

\subsection{Auxiliary Reward for Recovery From OOD Situations}
Because it is infeasible to directly calculate the distance of the agent's state from the learned state distribution, we approximate it using the uncertainty of the state predicted through MCD. When state $s_t$ is given, the uncertainty of $s_t$ is calculated by applying Eq. (\ref{eqn:mcdo_cal}) to the encoder network $e_\phi$, which is a component of policy $\pi_\phi$, as shown in Fig. \ref{fig:arch}. The calculated uncertainty ${\sigma}^u$ is a vector that has the same dimensions as the output because it approximates the variance of the output distribution of $e_\phi$. To represent the relative distance of the state from the learned state distribution using the uncertainty vector $\sigma^u$, we propose using the mapping function $g: \mathbb{R}^d \rightarrow \mathbb{R}$, which maps $\sigma^u$ to the uncertainty distance $d^u$. First, each element in $\sigma^u$ is normalized using min-max normalization to generalize the range of each element to between $[0,1]$; this enables the uncertainty distance to have a generalized range for all environments. The element-wise maximum value of $\sigma^u$ is saved in the policy and updated whenever a new maximum value occurs during the training as follows:
\begin{equation}
\footnotesize
    \langle \sigma^{max} \rangle_i=\begin{cases}
            -\infty & \textrm{if}\ t=0\\
            \langle \sigma^u_{t} \rangle_i\ & \textrm{else if}\ \langle \sigma^u_{t} \rangle_i > \langle \sigma^{max}\rangle_i\\
            \langle \sigma^{max} \rangle_i & \textrm{otherwise}
            \end{cases}, \forall i\in \{1,2...,d\},
\end{equation}
where ${\langle\cdot\rangle}_i$ is the $i^{\textrm{th}}$ element of the vector, $d$ is the dimension of the $\sigma^u$, and $t$ is the time step of the training. Because MCD approximates the variance of the output distribution of $e_\phi$, the element-wise minimum value of $\sigma^u$ is set to zero which is the lower bound of the variance. Finally, the uncertainty distance $d^u$ is calculated by normalizing each element of $\sigma^u$ and taking the element-wise weighted average of the normalized uncertainty vector as follows:
\begin{equation}
\footnotesize
    d^u_t = g(\sigma^{u}_t) = \sum_{i=1}^{d}w_{i}\frac{{\langle \sigma^{u}_t \rangle}_i}{{\langle\sigma^{max}\rangle}_i},\ \ 
    \textrm{where}\ \ w_{i} = \frac{\frac{\langle\sigma^{u}_t\rangle_i}{\langle\sigma^{max}\rangle_i}}{\sum_{i=1}^{d}\frac{\langle\sigma^{u}_t\rangle_i}{\langle\sigma^{max}\rangle_i}}.
\end{equation}
We used the weighted average to prevent an element with high uncertainty from being offset by one with low uncertainty. As each element of the uncertainty vector is guaranteed to be in the range of $[0, 1]$, the uncertainty distance is also in the range of $[0, 1]$. The uncertainty distance is close to 1 when the state of the agent has high uncertainty, which means the state is far from the learned state distribution. In contrast, the uncertainty distance is close to 0 when the state of the agent has low uncertainty, which means the state is close to the learned state distribution. We designed an auxiliary reward as a negative uncertainty distance as follows:
\begin{equation}
\footnotesize
r^u(s_t,a_t,s_{t+1})=-d^u_{t+1}=-g\big(\sigma^u_{t+1}\big).
\end{equation}
When the agent chooses action $a_t$ at state $s_t$ and reaches the next state $s_{t+1}$, it receives a reward according to the uncertainty distance of the next state. When the agent takes an action that minimizes the uncertainty distance of the next state, which means approaching the learned state distribution, it receives a high auxiliary reward. Accordingly, by maximizing the expected cumulative auxiliary reward, the agent can learn how to return to the learned state distribution.

\subsection{Uncertainty-Aware Policy Consolidation (UPC)}
Catastrophic forgetting for the original task is problematic in terms of the sample efficiency of the retraining phase because the agent should relearn the original task after learning how to return to the learned state distribution. In order to prevent the agent from forgetting the original task during the retraining phase, we proposed UPC loss as follows:
\begin{equation}
\footnotesize
    \mathcal{L}_{con}^\pi = (1-d_t^u)D_{KL}\left(\pi(a_t|s_t)||\pi_{{org}}(a_t|s_t)\right),
\end{equation}
where $\pi_{{org}}$ is the policy for the original task, which is the fixed policy after the training phase. When the uncertainty distance is small, which indicates that the agent is close to the learned state distribution, the effect of the UPC is increased to regularize the policy to take action that is similar to the original policy to solve the original task. In contrast, when the agent’s state is far from the learned state distribution, the effect of the UPC is reduced, which enables the agent to learn new behavior to return to the learned state distribution.

\subsection{Self-Supervised RL for Recovery From OOD Situations}
As SeRO expands SAC, we consider the parameterized Q-function $Q_\theta(s_t,a_t)$ and the policy $\pi_\phi(a_t|s_t)$. The policy and Q-function are updated based on soft value iteration using experience sampled from the replay buffer $\mathcal{D}$. The $Q_\theta$ is updated by minimizing Eq. (\ref{policy_eval}) where the reward function $r_t$ is defined as follows:
\begin{equation}
\label{eqn:critic_obj}
\footnotesize
    r_t =\begin{cases}
            r^e_t & \textrm{if}\ s_{t} \in \mathcal{S}_{in}\\
            \lambda r^u_t& \textrm{else if}\ s_{t} \in \mathcal{S}_{OOD} = (\mathcal{S}_{in})^c
        \end{cases}, 
\end{equation}
where $r^e$ is an environmental reward for the original tasks, $\lambda$ is the weight coefficient, and $\mathcal{S}_{in}$ and $\mathcal{S}_{OOD}$ correspond to in-distribution state space and OOD state space, respectively. The objective for updating the policy $\pi_\phi$ is based on Eq. (\ref{policy_improv}) and augmented by UPC loss to regularize the policy. The augmented objective for training $\pi_\phi$ is as follows:
\begin{equation}
\label{eqn:training_obj}
\footnotesize
J(\phi)=\mathbb{E}_{(s_t,d^u_t)\sim \mathcal{D},a_t\sim\pi_\phi}\left[\alpha\log\pi_{\phi}(a_t|s_t)-Q_\theta(s_t,a_t) + \mathcal{L}_{con}^{\pi_\phi} \right].
\end{equation}
By updating the Q-function and the policy alternately, the agent is trained to return to the learned state distribution in the OOD states using the proposed auxiliary reward, and trained to solve the original tasks using an environmental reward when the agent returns to the learned state distribution. A detailed explanation of the overall retraining procedure can be found in the supplementary material.

\section{Experiments}
\label{sec:experiments}
Our in-depth experiments were designed to answer the following questions: 1) Is retraining for recovery from OOD situations necessary? 2) Can the proposed uncertainty distance successfully represent the relative distance of the state from the learned state distribution? 3) Can our method improve the agent's ability to recover from OOD situations compared to the baseline? 4) Can our method self-recognize whether the agent is in an OOD state and retrain the agent to recover from OOD situations? 5) What is the effect of each component of the proposed method?

We conducted experiments on four OpenAI gym's MuJoCo environments \cite{https://doi.org/10.48550/arxiv.1606.01540} to answer the above questions. To evaluate the improvement of the agent’s ability to recover from OOD situations, we used SAC as a baseline because our method was implemented by expanding it. In the remainder of this section, we explain how we implemented the environments for the experiments and the result of the experiment for each question in detail. Note that the experiments for the second and fifth questions can be found in the supplementary material.

\renewcommand{\arraystretch}{1.7}
\begin{table*}[t]
\centering
\resizebox{\textwidth}{!}{
\begin{tabular}{|c||cc|cc|cc|cc|}
\hline
Method &
\multicolumn{2}{c||}{HalfCheetah-v2} & \multicolumn{2}{c||}{Hopper-v2} & \multicolumn{2}{c||}{Walker2D-v2} & \multicolumn{2}{c|}{Ant-v2}\\ \cline{2-9} 
& \multicolumn{1}{c|}{HalfCheetahNormal-v2} & \multicolumn{1}{c||}{HalfCheetahOOD-v2} & \multicolumn{1}{c|}{HopperNormal-v2} & \multicolumn{1}{c||}{HopperOOD-v2} & \multicolumn{1}{c|}{Walker2DNormal-v2} & \multicolumn{1}{c||}{Walker2DOOD-v2} & \multicolumn{1}{c|}{AntNormal-v2} & \multicolumn{1}{c|}{AntOOD-v2}  \\ \hhline{|=||=|=||=|=||=|=||=|=|}

SAC 
& \multicolumn{1}{c|}{11895.96$\pm$591.54} &  \multicolumn{1}{c||}{-635.38$\pm$112.47} 
& \multicolumn{1}{c|}{3411.02$\pm$146.90} & \multicolumn{1}{c||}{758.47$\pm$366.47}  
& \multicolumn{1}{c|}{4069.70$\pm$432.95} &  \multicolumn{1}{c||}{576.48$\pm$527.65} 
& \multicolumn{1}{c|}{5555.20$\pm$866.35} & \multicolumn{1}{c|}{0.86$\pm$8.49} \\ \hline 
\end{tabular}
}\\[-1.0ex]
\caption{Average returns computed over 100 episodes after the training phase.}
\label{table1}
\vspace{-1.5em}
\end{table*}

\subsection{Environments}
\begin{figure}[t]
    \centering
    \subfigure[Training environments]{\includegraphics[width=\linewidth]{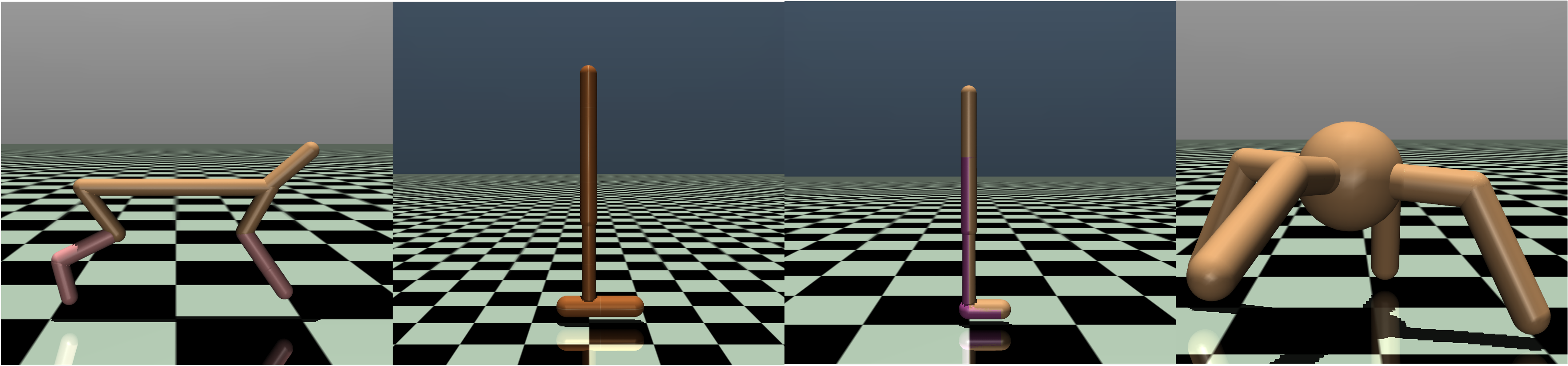}}\\[-1.0ex]
    \subfigure[Retraining environments]{\includegraphics[width=\linewidth]{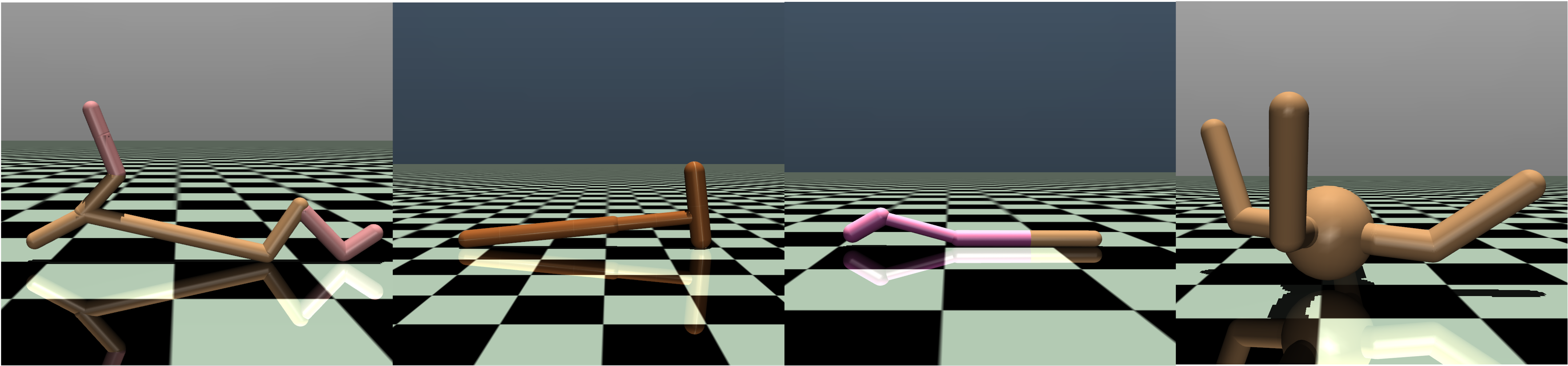}}\\[-2.0ex]
    \caption{Training environments (top) and retraining environments (bottom). From left: HalfCheetah-v2, Hopper-v2, Walker2D-v2, and Ant-v2.}
    \label{fig:envs}
    \vspace{-1.5em}
\end{figure}
\label{subsec:setup}
We used HalfCheetah-v2, Hopper-v2, Walker2D-v2, and Ant-v2 from the gym's MuJoCo environments. To evaluate our method, we modified the original environments to implement training environments (e.g., HalfCheetahNormal-v2, HopperNormal-v2, Walker2DNormal-v2, and AntNormal-v2) and retraining environments (e.g., HalfCheetahOOD-v2, HopperOOD-v2, Walker2DOOD-v2, and AntOOD-v2) separately as shown in Fig. \ref{fig:envs}. In the retraining environments, we emphasize that a trained agent is spawned in a state that is \textit{unseen in the training phase}. In the experiments, we denoted states in the training environments as the in-distribution states, and the states belonging to the rest of the state space as the OOD states for the sake of clarity. We want to note that our method solely addressed OOD states that can be recovered by the agent based on its physical characteristics.

\noindent \textbf{HalfCheetah-v2}: In HalfCheetahNormal-v2, the episode is terminated when the agent flips over. Whether the agent flips over is determined by whether the angle of the agent's front tip is out of a certain range. In HalfCheetahOOD-v2, the agent is spawned upside down. To return to the learned state distribution, the agent should learn how to \textit{turn its body over}.

\noindent \textbf{Hopper-v2 \& Walker2D-v2}: In HopperNormal-v2 and Walker2DNormal-v2, the episode is terminated when the agent falls down. Whether the agent falls down is determined by whether the angle of the agent's top part or the height is out of a certain range. The agent is spawned lying face up and lying face down on the floor in HopperOOD-v2 and Walker2DOOD-v2, respectively. To return to learned state distribution, the agent should learn how to \textit{stand up}.

\noindent \textbf{Ant-v2}: In AntNormal-v2, the episode is terminated when the agent flips over. Whether the agent flips over is determined when the pitch or roll angle of the agent's torso is out of a certain range. In AntOOD-v2, the agent is spawned upside down. To return to the learned state distribution, the agent should learn how to \textit{turn its body over}.

We refer the readers to the supplementary material for a more detailed explanation of the environments.

%===============================================================================
\subsection{Analysis of the Necessity of Retraining}
In this subsection, we analyze whether the retraining of the trained agent is necessary for OOD situations. If agents trained for solving the original tasks can perform well in the OOD states without retraining, retraining for recovery from OOD situations may not be needed. To verify the necessity of retraining, we first trained the agents in the training environments for 1 million steps using SAC, and we then evaluated the trained agents in the training environments and retraining environments respectively. Table \ref{table1} displays the average returns of the environmental reward for the original tasks computed over 100 episodes on five random seeds. As shown in the table, the average return of the agent was significantly lower in the retraining environments than in the training environments. This result suggests that the trained agent could not perform the original tasks well when it fell into an OOD state, and therefore, retraining for recovery from OOD situations is necessary.

\begin{figure*}[t]
    \centering
    \subfigure[HalfCheetah-v2]{\includegraphics[width=0.245\linewidth]{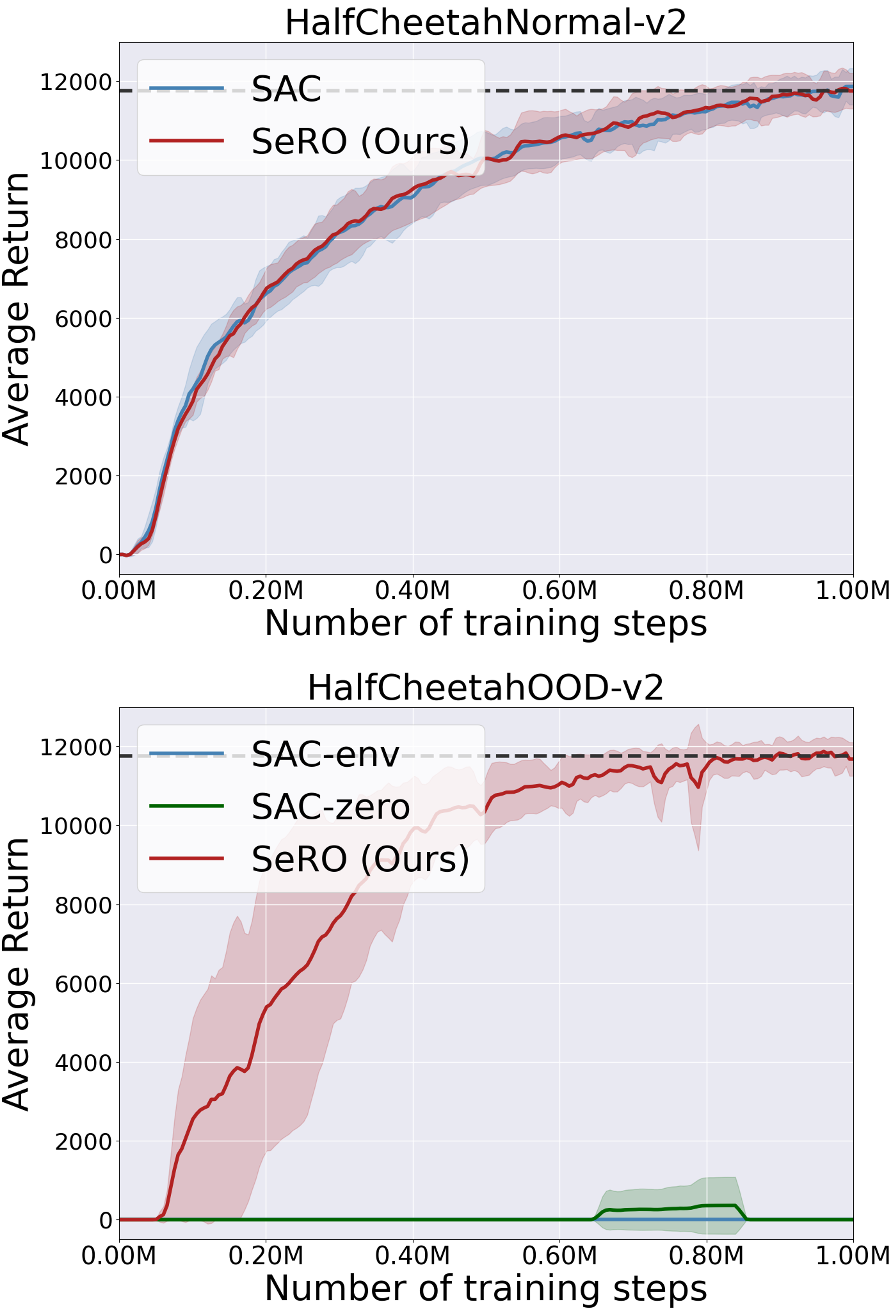}}
    \subfigure[Hopper-v2]{\includegraphics[width=0.245\linewidth]{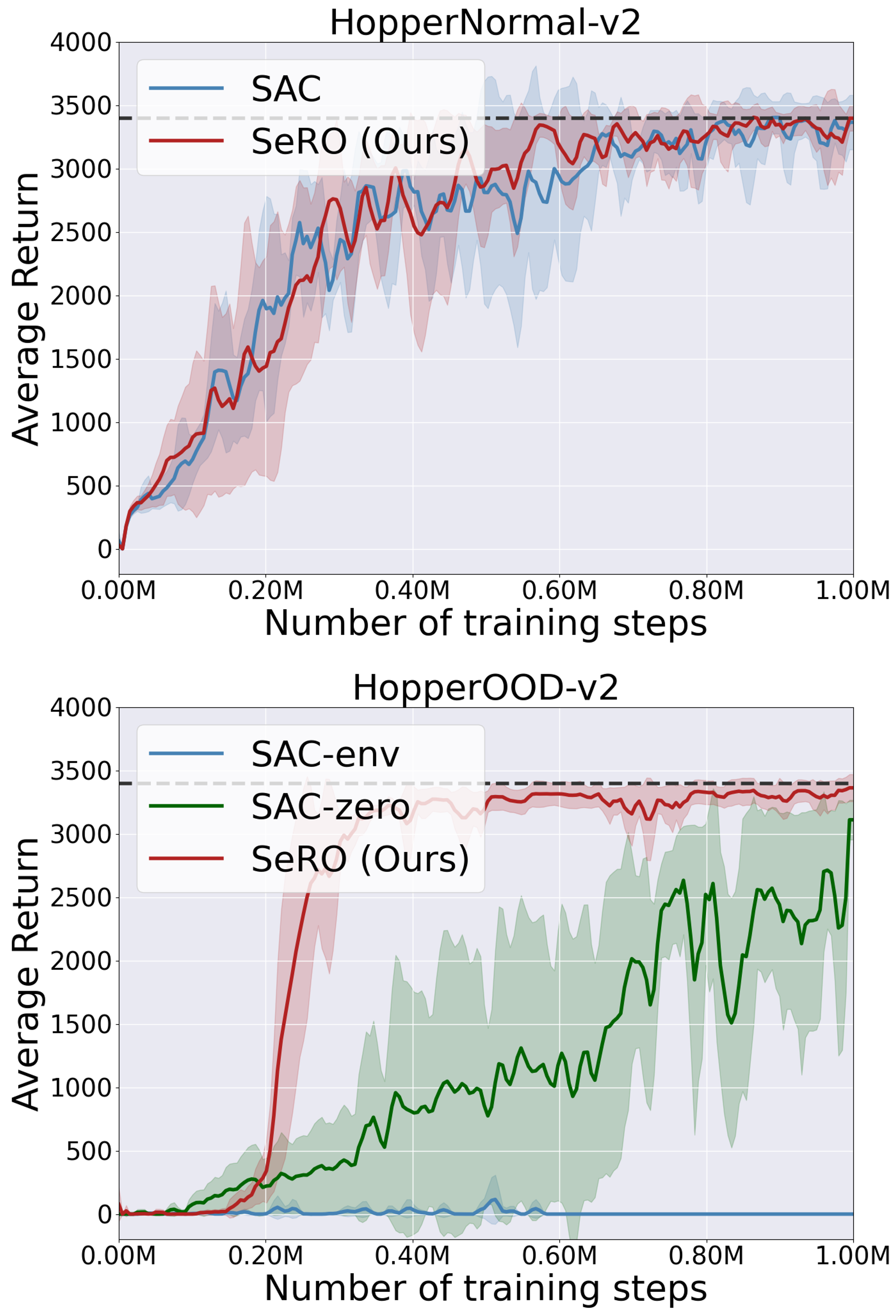}}
    \subfigure[Walker2D-v2]{\includegraphics[width=0.245\linewidth]{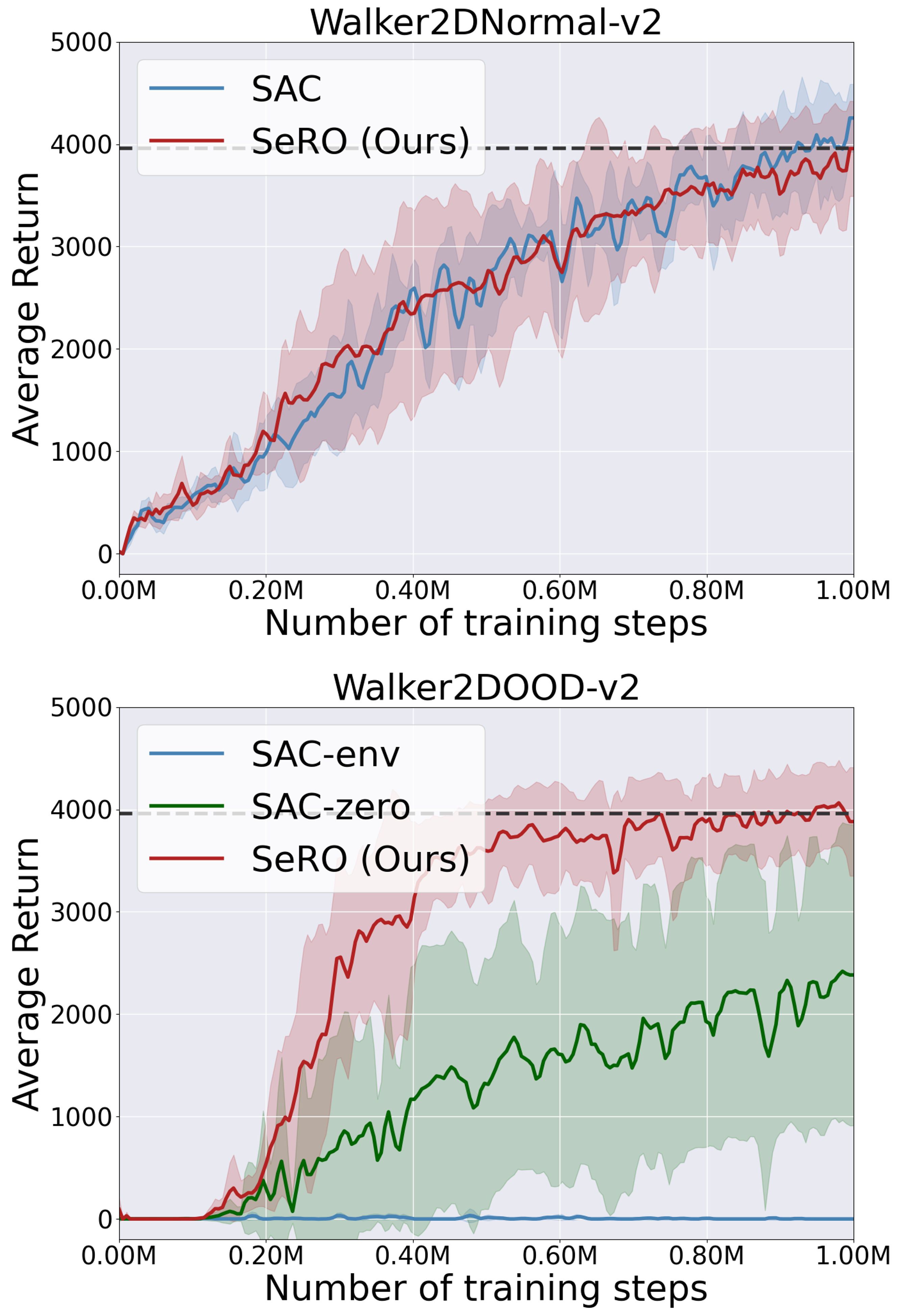}}
    \subfigure[Ant-v2]{\includegraphics[width=0.245\linewidth]{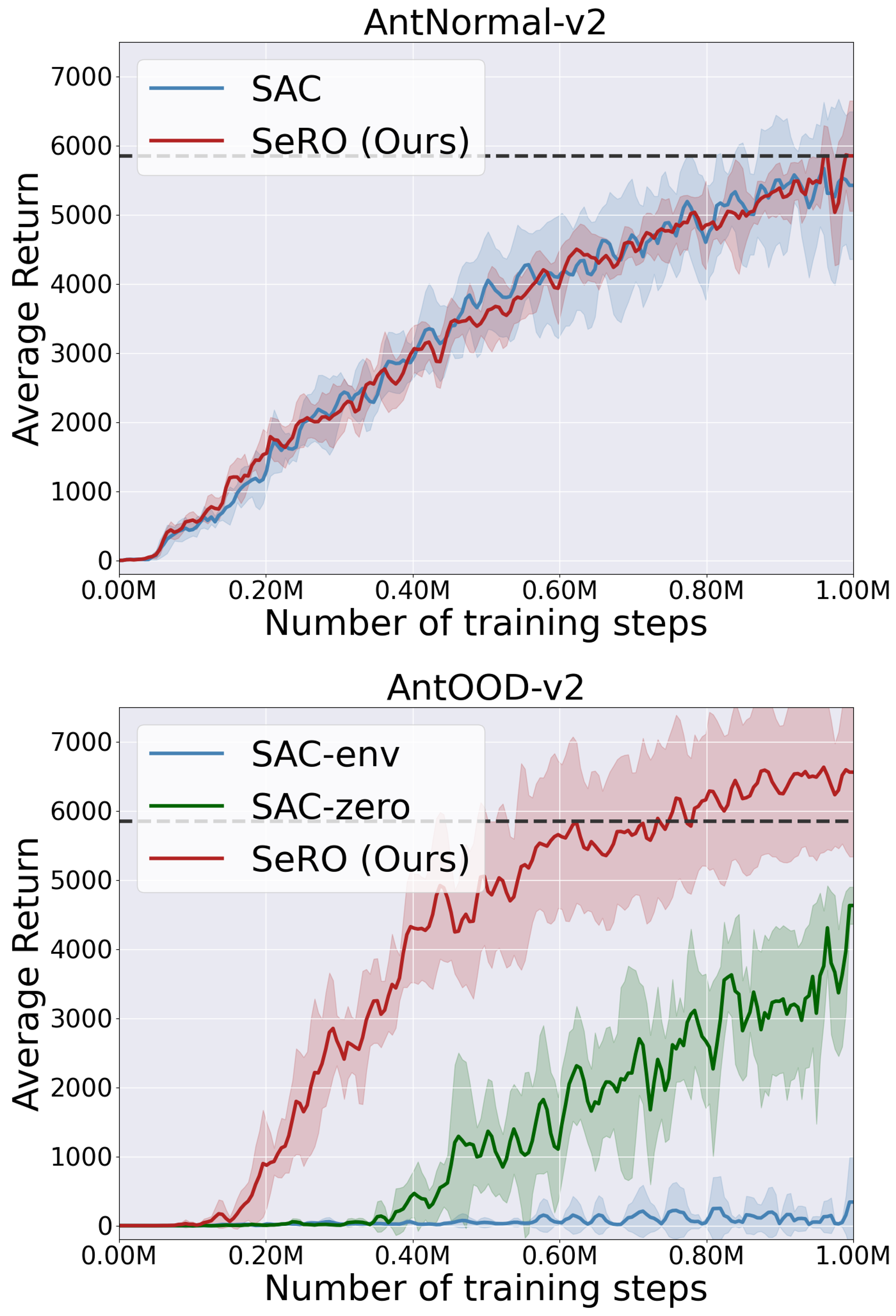}}\\[-2ex]
    \caption{Learning curves for the training environments (top) and the retraining environments (bottom), calculated for five episodes of evaluation at every 5000 steps of training. The darker-colored lines and shaded areas represent the average returns and standard deviations computed over five random seeds. The black dashed horizontal line represents the average return of the SeRO agent after the training phase.}
    \vspace{-1.2em}
    \label{fig:quantative}
\end{figure*}

\begin{figure*}[t]
    \centering
    \includegraphics[width=0.95\linewidth]{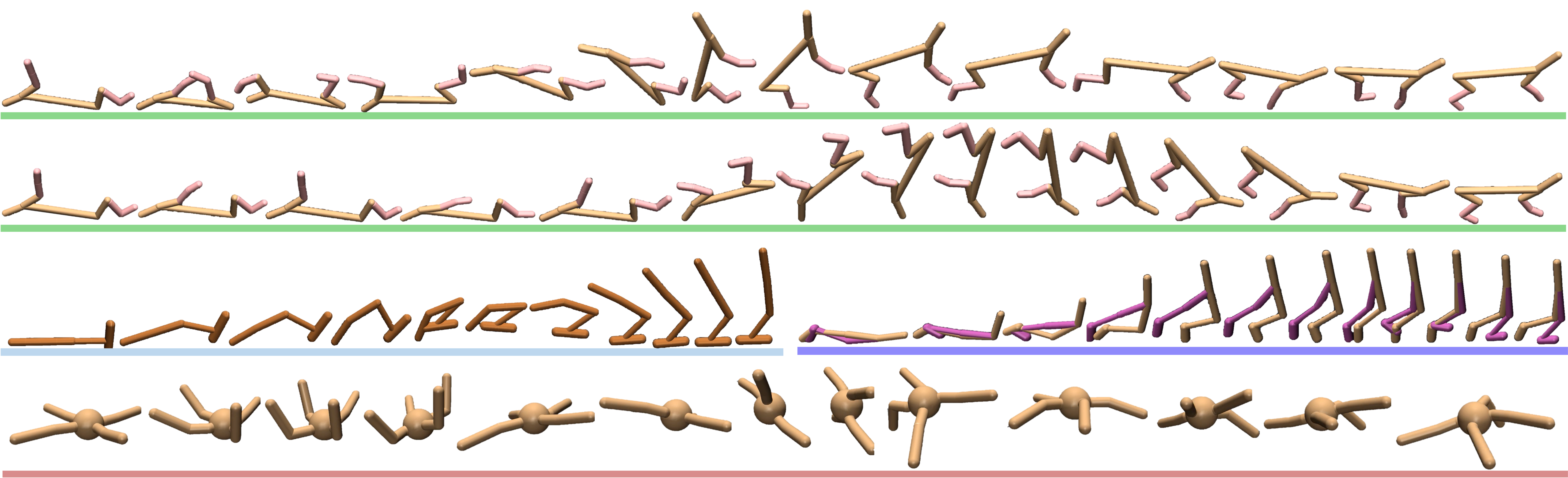}
    \caption{Qualitative results of SeRO after the retraining phase. The figure shows the agent's motion for a time increasing from left to right in each environment.}
    \label{fig:qualitative}
    \vspace{-1.4em}
\end{figure*}

\subsection{Retraining for Recovery From OOD Situations}
\label{subsec:sec_44}
We conducted an experiment to compare our method and SAC to answer the question that whether our method can improve the agent's ability to recover from OOD situations. Both methods are trained in the training phase for 1 million steps. Subsequently, the trained agents are retrained to recover from OOD situations in the retraining phase. We retrained the SAC agent in two different ways; One is SAC--env which receives \textit{environmental rewards} for the original tasks in OOD states, and the other is SAC--zero which receives \textit{zero rewards} in OOD states. Both methods receive environmental rewards for the original tasks in in-distribution states. In the case of SAC--zero, the agent is guided to return to the learned state distribution because the agent has a chance to receive a high reward if it visits in-distribution states while receiving zero rewards if it stays in OOD states. Fig. \ref{fig:quantative} shows learning curves acquired in the training phase and retraining phase for each environment. The learning curves represent the average return of the environmental reward for the original tasks. We emphasize that the learning curves for the retraining phase are calculated by setting the reward in OOD states to \textit{zero} to intuitively represent whether the agent successfully learns to return to in-distribution states so that a higher average return means that the agent returned to the learned state distribution and performed the original tasks successfully.

As shown in the figure, both methods exhibited comparable performance in the training environments. However, in the retraining phase, SAC--env agent failed to recover from OOD situations in all environments. We found that the agent falls into a local optimum and tries to perform the original tasks without returning to the learned state distribution, e.g., in AntOOD-v2, the agent moves without turning their body. Further analysis and qualitative results of the SAC--env after the retraining phase can be found in the supplementary material. When comparing SeRO and SAC--zero, SeRO showed much higher sample efficiency and average return than SAC--zero. Moreover, while SAC--zero failed to restore the original performance, SeRO successfully restored the original performance for all environments considering that the average return converges to that reached in the training phase. These results are due to the two characteristics of SeRO. First, the auxiliary reward enables SeRO agents to distinguish the values of OOD states according to their distance from the learned state distribution even before visiting in-distribution states. We refer the readers to the experiment for the second question in the supplementary material for whether the uncertainty distance can approximate the distance of the state from the learned state distribution. In contrast, SAC--zero agents cannot distinguish the values of OOD states until it visits in-distribution states because it receives constant zero rewards in OOD states. Therefore, SeRO agents can quickly learn how to return to learned state distribution, while SAC--zero agents cannot learn until they first visit in-distribution states through exploration. We found that SAC--zero failed to return to the learned state distribution in HalfCheetahOOD-v2 where in-distribution states are difficult to visit through exploration, while our method succeeded for all seeds. Second, UPC prevents SeRO agents from forgetting the original tasks during learning to return to the learned state distribution and also regularizes agents to take action similar to what they learned in the training phase once they return to in-distribution states. Therefore, once the agent learns to return to the learned state distribution, SeRO converges quickly to the average return comparable to that reached in the training phase. To summarize the results, the reward for the original tasks cannot lead the agent to return to the learned state distribution and therefore the reward for recovery should be defined separately. The constant zero rewards in OOD states are also not suitable because they cannot guarantee the agent to return to the learned state distribution in environments where in-distribution states are difficult to visit through exploration. In contrast, the proposed auxiliary reward successfully leads the agent to return even in such environments. Moreover, the proposed UPC enables the agent to restore the original performance for the original tasks quickly by preventing catastrophic forgetting. As a result, our method improves the agent’s ability to recover from OOD situations in terms of sample efficiency and restoration of the original performance for the original tasks. 

Fig. \ref{fig:qualitative} visualizes the qualitative results of the agent trained using SeRO after the retraining phase. In HalfCheetahOOD-v2, the agent first shakes its legs to create rotational force and then hits the floor and tumbles to turn over. As shown in the figure, because our auxiliary reward is not designed explicitly for the desired behavior, agents learn diverse ways to return to the learned state distribution, such as tumbling forward or backward. In HopperOOD-v2, the agent first bends its top part to raise the body and then stands up by pushing the floor using the top part. In Walker2DOOD-v2, the agent first raises its top part, steps on the floor with one knee, and then stands up by pushing the floor with the knee. In AntOOD-v2, the agent turns its body over with momentum by hitting the floor with its legs. Although there is no explicit reward based on the prior knowledge of the OOD situations and environments, the agent trained using SeRO successfully learns the desired behavior to return to the learned state distribution in a self-supervised manner.

\subsection{Retraining for Recovery From OOD Situations With the Agent’s Own Criterion}
\begin{figure}[t]
    \centering
    \subfigure[HalfCheetahOOD-v2]{\includegraphics[width=0.48\linewidth]{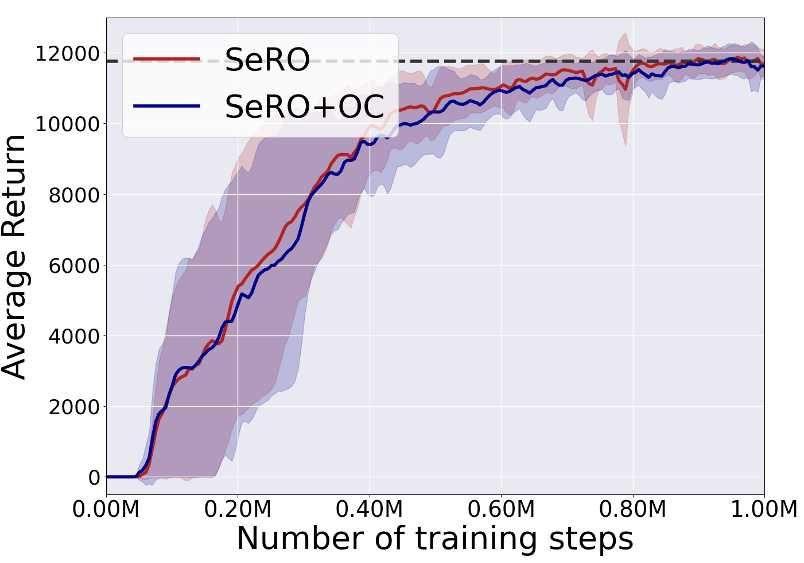}}
    % \hspace{0.05\linewidth}
    \subfigure[HopperOOD-v2]{\includegraphics[width=0.48\linewidth]{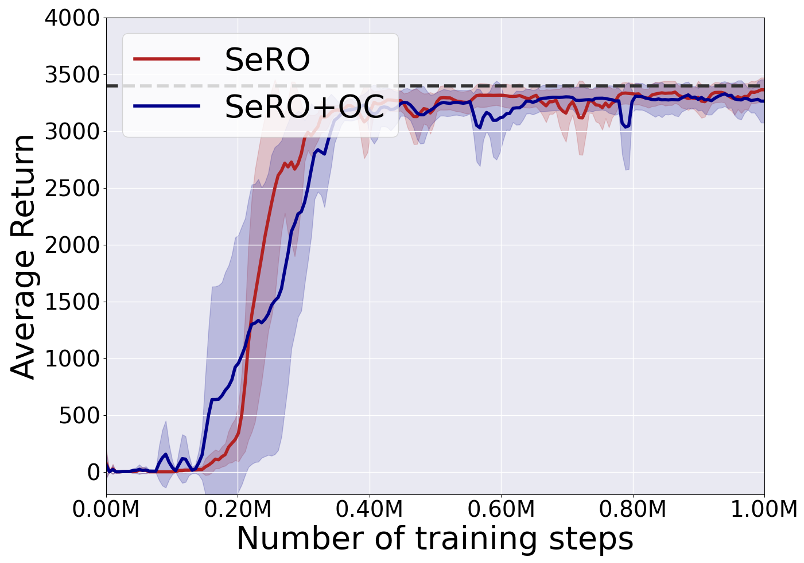}}\\[-2ex]
    \subfigure[Walker2DOOD-v2]{\includegraphics[width=0.48\linewidth]{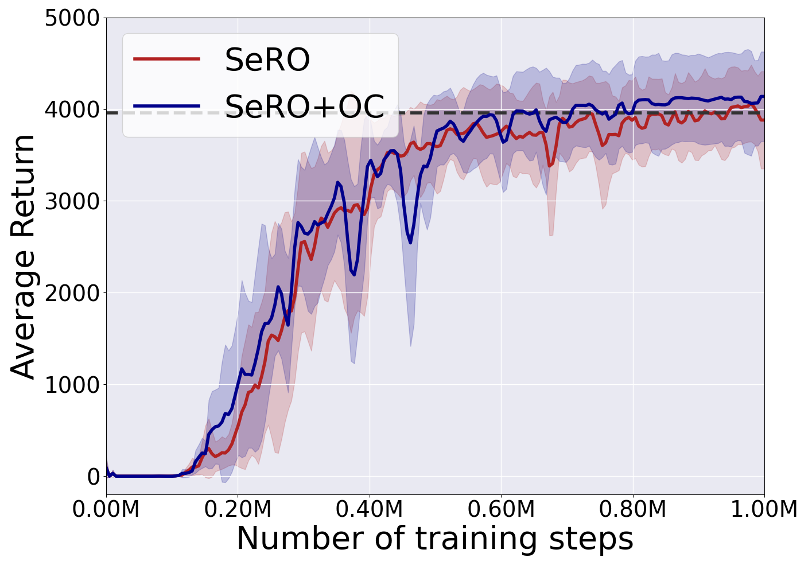}}
    % \hspace{0.05\linewidth}
    \subfigure[AntOOD-v2]{\includegraphics[width=0.48\linewidth]{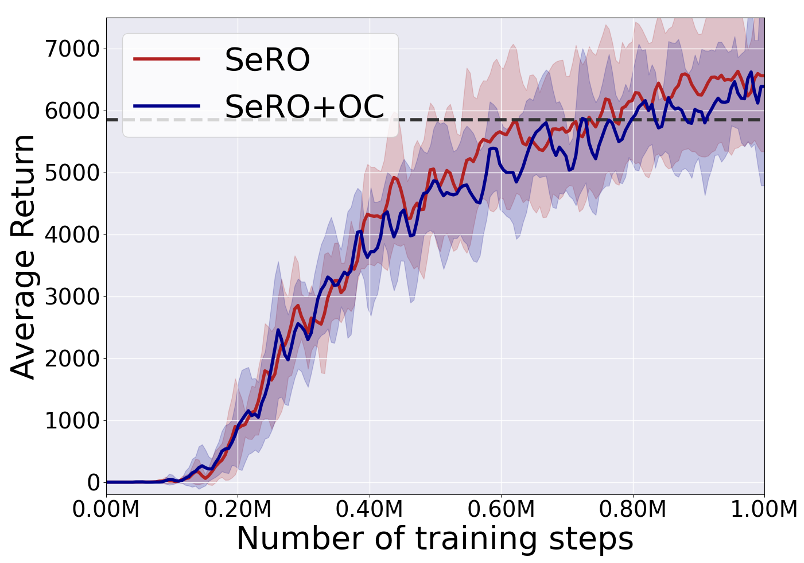}}\\[-2ex]
    
    \caption{Learning curves for the retraining environments of SeRO and SeRO+OC, calculated for five episodes of evaluation at every 5000 steps of training. The darker-colored lines and shaded areas represent the average returns and standard deviations computed over five random seeds. The black dashed horizontal line represents the average return of the SeRO agent after the training phase.}
    \vspace{-1.1em}
    \label{fig:own_crit}
\end{figure}
In the previous subsections, we defined the states in the training environments as in-distribution states $s_{in}\in \mathcal{S}_{training}$ and states belonging to the rest of the state space as the OOD states $s_{OOD}\in (\mathcal{S}_{training})^c$ for the experiments. Accordingly, the reward function for retraining SeRO in Section \ref{subsec:sec_44} can be formulated as follows:
\begin{equation}
\label{eqn:manual_crit}
\footnotesize
    r_t =\begin{cases}
            r^e_t & \textrm{if}\ s_{t} \in \mathcal{S}_{training}\\
            \lambda r^u_t& \textrm{otherwise}
        \end{cases}, 
\end{equation}
However, determining whether the current state is in $\mathcal{S}_{training}$ requires prior knowledge of $\mathcal{S}_{training}$, which is infeasible in real-world environments. The reason we used explicit criterion for distinguishing in-distribution states and OOD states in this way was for a fair comparison with the baseline (SAC--zero), which could not recognize whether the current state is the in-distribution state or the OOD state itself. To retrain the agent completely without prior knowledge or human intervention, the agent should recognize whether the current state is the OOD state or the in-distribution state based on its own criterion. In this experiment, we retrained SeRO with its own criterion based on the uncertainty distance. The corresponding reward function for retraining SeRO can be formulated as follows:
\begin{equation}
\label{eqn:own_crit}
\footnotesize
    r_t =\begin{cases}
            r^e_t & \textrm{if}\ d_{t}^u<\epsilon\\
            \lambda r^u_t& \textrm{otherwise}
        \end{cases}, 
\end{equation}
where $\epsilon$ is the threshold. When the uncertainty distance is smaller than $\epsilon$, which means that the agent's state is close to the learned state distribution, the agent is trained using environmental rewards to solve the original task. Otherwise, the agent is trained using auxiliary rewards to return to the learned state distribution.

We first trained the SeRO agent in the training environments for 1 million steps. Subsequently, the trained agent is retrained in the retraining environments with the reward function Eq. (\ref{eqn:own_crit}). Fig. \ref{fig:own_crit} shows the learning curves of the SeRO trained in Section \ref{subsec:sec_44} and the SeRO trained with its own criterion which we refer to as SeRO+OC. Note that the same trained agent is used for both algorithms in the retraining phase. For a fair comparison with SeRO, the average return of SeRO+OC in the plots is calculated in the same way as in Section \ref{subsec:sec_44} using an explicit criterion based on the prior knowledge of $\mathcal{S}_{training}$, although SeRO+OC is actually trained using the reward function in Eq. (\ref{eqn:own_crit}) with its own criterion. As shown in the figure, SeRO+OC shows comparable performance to SeRO for all environments. The results demonstrate that our method can learn to return to the learned state distribution and perform original tasks successfully with its own criterion by self-recognizing whether the current state is the OOD state or the in-distribution state according to the uncertainty distance, which enables self-supervised recovery without prior knowledge about environments and OOD situations.

\section{Conclusion}
\label{sec:limitations}
In this study, we addressed the situation in which a trained agent has already fallen into an OOD state, which can happen unintentionally in real-world environments. We introduced SeRO, a self-supervised RL method for recovery from OOD situations. In particular, we proposed an auxiliary reward that guides the agent to the learned state distribution. We also proposed uncertainty-aware policy consolidation to prevent the agent from forgetting the original task while learning how to return to the learned state distribution. We evaluated our method on OpenAI gym's MuJoCo environments and demonstrated that the proposed method can improve the agent's ability to recover from OOD situations. Moreover, we also showed that our method successfully learns to recover from OOD situations even when in-distribution states are difficult to visit through exploration.

\section*{Acknowledgments}
This work was supported by the MOTIE (Ministry of Trade, Industry, and Energy) in Korea, under the Human Resource Development Program for Industrial Innovation (P0017304) supervised by the Korea Institute for Advancement of Technology (KIAT), and in part by the National Research Foundation of Korea (NRF) through the Ministry of Science and ICT under Grant  2021R1A2C1093957. Cho was majoring in the Integrated Major in Smart City Global Convergence, Seoul National University when he was participating in this research. The Institute of Engineering Research at Seoul National University provided research facilities for this work.

%% The file named.bst is a bibliography style file for BibTeX 0.99c
\bibliographystyle{named}
\bibliography{ijcai23}

\end{document}